\documentclass[runningheads]{llncs}

\usepackage{eccv}

\usepackage{eccvabbrv}

\usepackage{graphicx}
\usepackage{booktabs}
\usepackage{wrapfig}
\usepackage{amsmath}
\usepackage{amsfonts}
\usepackage{amssymb}
\usepackage{dsfont}
\usepackage{multirow}
\usepackage{comment}

\usepackage[accsupp]{axessibility}  % Improves PDF readability for those with disabilities.

\usepackage{hyperref}

\usepackage{orcidlink}

\usepackage{times}
\usepackage{soul}
\usepackage{url}
\begin{document}

\title{Fairness-aware Vision Transformer via Debiased Self-Attention}

% \titlerunning{F}

\author{Yao Qiang \orcidlink{0000-0003-2995-3385} \and
Chengyin Li\orcidlink{0000-0003-2450-9760} \and
Prashant Khanduri\orcidlink{0000-0003-3055-2917} \and
Dongxiao Zhu\orcidlink{0000-0002-0225-7817}}

\authorrunning{Y. Qiang et al.}

\institute{Wayne State University, Detroit MI 48202, USA \\
\email{\{yao,cyli,khanduri.prashant,dzhu\}@wayne.edu}}

\maketitle

\begin{abstract}
    Vision Transformer (ViT) has recently gained significant attention in solving computer vision (CV) problems due to its capability of extracting informative features and modeling long-range dependencies through the attention mechanism. Whereas recent works have explored the trustworthiness of ViT, including its robustness and explainability, the issue of fairness has not yet been adequately addressed. We establish that the existing fairness-aware algorithms designed for CNNs do not perform well on ViT, which highlights the need to develop our novel framework via Debiased Self-Attention (DSA). DSA is a fairness-through-blindness approach that enforces ViT to eliminate spurious features correlated with the sensitive label for bias mitigation and simultaneously retain real features for target prediction. Notably, DSA leverages adversarial examples to locate and mask the spurious features in the input image patches with an additional attention weights alignment regularizer in the training objective to encourage learning real features for target prediction. Importantly, our DSA framework leads to improved fairness guarantees over prior works on multiple prediction tasks without compromising target prediction performance. Code is available at \href{https://github.com/qiangyao1988/DSA}{https://github.com/qiangyao1988/DSA}. 
\end{abstract}

\section{Introduction}

\noindent
Vision Transformer (ViT) \cite{dosovitskiy2020image} has emerged as an architectural paradigm and a viable alternative to the standard Convolutional Neural Network (CNN). ViT extracts global relationships via the attention mechanism leading to impressive feature representation capabilities, resulting in improved performance in various CV tasks, such as image classification \cite{liu2021swin}, object detection \cite{carion2020end,dai2021up}, semantic segmentation \cite{strudel2021segmenter,li2023focalunetr}, and image generation \cite{hudson2021generative}, to name a few.

Given its promising performance, researchers have studied the trustworthiness of ViT for real-world applications. Several works \cite{mao2022towards,zhou2022understanding,chefer2021transformer,qiang2022attcat,li2023negative,qiang2023interpretability} have explored the robustness and explainability of ViT, highlighting its strengths and weaknesses. Robustness refers to the ability of ViT to perform well on inputs that deviate from the training distribution, such as adversarial examples \cite{mao2022towards,zhou2022understanding,qiang2024prompt}. Explainability refers to the ability of ViT to provide insights into its decision-making process, which is crucial for building trust in the model \cite{chefer2021transformer,qiang2022attcat,li2023negative,qiang2023interpretability}.

\begin{figure}[t]
    \centering
    \includegraphics[width = 0.5\linewidth]{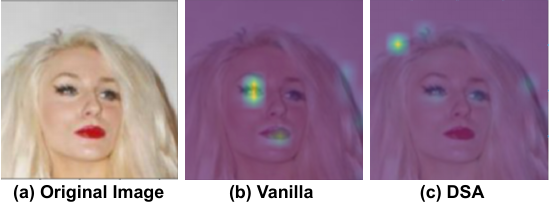}
    \caption{An illustration example. The prediction target label is \emph{Hair Color} and the sensitive label is \emph{Gender}. The heatmap of attention weights shows that the Vanilla ViT uses spurious features, e.g., `red lip' and `eye shadow', whereas the fairness-aware ViT via our DSA leverages the real features, e.g., `hair', for target prediction. \label{fig:illustration}}
\end{figure}

Besides robustness and explainability, fairness stands as another core trustworthy desiderata \cite{holstein2019improving,chouldechova2018frontiers,qiang2022counterfactual}. Several studies have already demonstrated that many deep-learning-based models simply make predictions by exploiting spurious correlations present in the training data \cite{wang2020mitigating,jung2021fair}. These spurious correlations occur when a feature is statistically informative for a majority of training examples but do not capture the underlying relationship between certain input features and the target outputs \cite{wilson2019predictive,singh2020don}.

Specifically, \textbf{real features} are genuinely correlated with the target outputs and can be generalized to other data sets. Whereas \textbf{spurious features}, also called short-cut features, are spuriously correlated with the target outputs in certain data sets only and do not generalize. As the example shown in Figure \ref{fig:illustration}, spurious features like `eye shadow' or `red lips' are spuriously correlated with hair color in the training data set. The vanilla ViT simply learns these spurious features as a shortcut to predict the hair color rather than learning the real features that are relevant to the target label, as shown in Figure \ref{fig:illustration}(b). This is where the fairness-aware ViT comes in - it is designed to learn real features that are not correlated with the sensitive label, in this case, gender, to make unbiased predictions as shown in Figure \ref{fig:illustration}(c).

Although an array of debiasing algorithms have been proposed \cite{madras2018learning,zhang2018mitigating,zhang2020towards,jung2021fair,wang2022fairness,qiang2022counterfactual} for CV tasks, most are designed for learning with the CNN models. Whether these algorithms are compatible or even transferable to the ViT architecture is still an open research question. Regardless of the model architecture, limiting the spurious correlation between the input features and the target outputs for bias mitigation is still a challenging problem. One key challenge arises from the fact that automatically locating the spurious features in the input images is computationally intractable. For example, one simple solution is to have domain experts and/or crowd workers curate the entire training set, which neither works well with unknown bias \cite{li2022discover} nor is scalable to large-scale datasets \cite{mcdonnell2016relevant}. Moreover, even if one can identify the spurious features, another major challenge is how to make the classifier blind to such features. Image in-painting \cite{wang2022dual}, can be a solution but has limitations, such as dependency on algorithm quality and feature selection accuracy, and difficulty in balancing feature removal with image integrity.

To address the above challenges, we propose a novel framework for ensuring bias mitigation training of ViT via Debiasing Self-Attention (DSA). DSA uses a hierarchical method. First, it localizes and perturbs spurious features in image patches by using adversarial attacks on a bias-only model. This model is trained to predict sensitive labels like gender and race by maximizing the use of spurious features and minimizing the use of real features like hair color. Specifically, we design a novel adversarial attack against the bias-only model to capture the most important patches for learning spurious features. Different form image in-painting approaches, adversarial examples are automatically constructed during the attack by directly perturbing the patches with spurious features. Second, the original training set is augmented with the constructed adversarial examples to formulate a debiased training set. In addition, a regularizer, which is specifically designed for the self-attention mechanism in ViT, is introduced to align the biased examples and their corresponding unbiased adversarial examples through attention weights. Our novel training objective encourages ViT models to learn real features while ensuring fairness.

% Adversarial attacks, although initially designed to evaluate the robustness of ViT, can also be used to identify and remove spurious features to train debiased ViTs.

%\noindent
We summarize our major contributions: (1) We design a novel DSA framework for ViT to mitigate bias in both the training set and learning algorithm. (2) We tackle several challenges for the under-addressed fairness problem in ViT from a novel perspective of leveraging adversarial examples to eliminate spurious features while utilizing attention weights alignment to retain real features. (3) The quantitative experimental results demonstrate that DSA improves group fairness while maintaining competitive or even better prediction accuracy compared to baselines. Our qualitative analysis further indicates that DSA has reduced attention to spurious features.

\section{Related Work}

\subsection{ViT for Image Classification}

\noindent
ViT has been a topic of active research since its introduction, and various approaches have been proposed to improve its performance and applicability. The earlier exploration of ViT either used a hybrid architecture combining convolution and self-attention \cite{carion2020end} or a pure self-attention architecture without convolution \cite{ramachandran2019stand}. The work in \cite{dosovitskiy2020image} proposed a ViT that achieves impressive results on image classification using an ImageNet dataset. This success has motivated a series of subsequent works to further exploit ViT's expressive power from various perspectives, such as incorporating locality into ViT \cite{li2021localvit,liu2021swin,yang2021focal}, and finding well-performing ViT using neural architecture search \cite{chen2021autoformer}. 

\subsection{Fairness and Debiased Learning}

\noindent
The existing techniques for fairness and debiased learning can be roughly categorized into pre-, in-, and post-processing. 

\noindent
\textbf{Pre-processing} methods attempt to debias and increase the quality of the training set with the assumption that fair training sets would result in fair models \cite{zhang2020towards,kim2021biaswap,chuang2021fair}. The work in \cite{zhang2020towards} proposed to balance the data distribution over different protected attributes by generating adversarial examples to supplement the training dataset. Similarly, \cite{kim2021biaswap} generated the bias-swapped image augmentations to balance protected attributes, which would remove the spurious correlation between the target label and protected attributes. In \cite{chuang2021fair}, the authors presented fair mixup as a new data augmentation method to generate interpolated samples to find middle-ground representation for different protected groups. The work \cite{qiang2022counterfactual} described a novel generative data augmentation approach to create counterfactual samples that d-separates the spurious features and the targets ensuring fairness and attribution-based explainability.

\noindent
\textbf{In-processing} approaches aim to mitigate bias during the training process by directly modifying the learning algorithm and model weights with specifically designed fairness penalties/constraints or adversarial mechanism \cite{madras2018learning,zhang2018mitigating,sagawa2019distributionally,kim2019learning,nam2020learning}. To enforce the fairness constraints, one line of works either disentangles the association between model predictions and the spurious features via an auxiliary regularization term \cite{nam2020learning} or minimizes the performance difference between protected groups with a novel objective function \cite{sagawa2019distributionally}. However, the issue is that the trained models may behave differently at the inference stage even though such fairness constraints are satisfied during the training. Another line of works \cite{madras2018learning,zhang2018mitigating,kim2019learning,wang2022fairness} enforce the model to generate fair outputs with adversarial training techniques through the min-max objective: maximizing accuracy while minimizing the ability of a discriminator to predict the protected (sensitive) attribute. Nevertheless, this process can compromise the model performance on the main prediction task. Additional lines of works impose either orthogonality \cite{sarhan2020fairness}, disentanglement \cite{locatello2019fairness}, or feature alignment \cite{jung2021fair} constraints on the feature representation and force the representation to be agnostic to the sensitive label. We note that most of these approaches are exclusively designed for CNN architectures, and whether these approaches are transferable to the ViT has not yet been demonstrated.

\noindent
\textbf{Post-processing} techniques directly calibrate or modify the classifier's decisions to certain fairness criteria at inference time \cite{kim2019multiaccuracy,lohia2019bias,alabdulmohsin2021near}. These methods require access to the sensitive attribute for fair inference, which may not be feasible in real-world applications due to salient security and privacy concerns.

\subsection{Fairness in ViT} 

\noindent
Recently, \cite{ghosal2022vision} explored how the spurious correlations are manifested in ViT and performed extensive experiments to understand the role of the self-attention mechanism in debiased learning of ViT. Despite the new insights, the authors did not provide any debiasing techniques for ViT. The authors in \cite{sudhakar2023mitigating} proposed a new method, named TADeT, for debiasing ViT that aims to discover and remove bias primarily from query matrix features. To our knowledge, this is the only published work along the line of fairness ViT. Nevertheless, this pioneering work TADeT has two weaknesses: first, it requires parameter sharing across the key and value weights in the self-attention mechanism, which may conflict with most ViT architectures; second, the complex alignment strategy on the query matrix is not straightforward, and well investigated. Thus, TADeT does not even outperform the compared baselines that are primarily designed for CNNs. 

In contrast to the above works, this work tackles the debiasing problem through a novel perspective of fairness-through-adversarial-attack. The proposed DSA framework combines the strengths of both pre- and in-processing approaches via leveraging data augmentation (for ensuring fairness in the training set) and feature alignment for bias mitigation. The adversarial examples are used to both disentangle spurious features from real features and to align attention weights, specifically, tailor-made for the self-attention mechanism in ViT. Notably, our approach for the fair ViTs is a novel addition to the growing body of work on ``adversarial examples for fairness" \cite{zhang2020towards,xu2021robust}. 

\section{Preliminaries}

\textbf{Overview of Vision Transformer:}
Similar to the Transformer architecture \cite{vaswani2017attention}, the ViT model expects the input to be a linear sequence of token/patch embeddings. An input image is first partitioned into non-overlapping fixed-size square patches resulting in a sequence of flattened 2D patches. These patches are then mapped to constant-size embeddings using a trainable linear projection. Next, position embeddings are added to the patch embeddings to imbibe relative positional information of the patches. Finally. the ViT model prepends a learnable embedding (class token) to the sequence of embedded patches following \cite{devlin2018bert}, which is used as image representation at the model's output.
To summarize, ViT consists of multiple stacked encoder blocks, where each block consists of a Multi-head Self Attention (MSA) layer and a Feed-Forward Network (FFN) layer. The MSA performs self-attention over input patches to learn relationships between each pair of patches, while the FFN layer processes each output from the MSA layer individually using two linear layers with a GeLU activation function. Both MSA and FFN layers are connected by residual connections.

\noindent
\textbf{Fairness Metrics:}
Many different notions of fairness have been proposed in the literature. In this work, we mainly focus on the two most widely used definitions: demographic parity and equalized odds as the metrics to assess the group fairness of the model \cite{hardt2016equality}. Demographic Parity (DP) compares the positive rates between all groups (defined by a sensitive label $s$, e.g., gender), particularly between the vulnerable minority group ($s=0$) and others ($s=1$), formally: $\mathrm{DP} = |\mathrm{PR}_{s=1} - \mathrm{PR}_{s=0}|$, where PR denotes the positive rate. Equalized Odds (EO) is used to understand the disparities in both the true positive rates and the false positive rates in the vulnerable group compared to others: $\mathrm{EO} = \frac{1}{2}|\mathrm{TPR}_{s=1} - \mathrm{TPR}_{s=0}| + \frac{1}{2}|\mathrm{FPR}_{s=1} - \mathrm{FPR}_{s=0}|$, TPR and FPR here represent true positive rate and false positive rate, respectively. In addition, we also use Accuracy (ACC) and Balanced Accuracy (BA) \cite{park2020readme}, where $\mathrm{BA} = \frac{1}{4}(\mathrm{TPR}_{s=0} + \mathrm{TNR}_{s=0} + \mathrm{TPR}_{s=1} + \mathrm{TNR}_{s=1})$, to evaluate the utility of the model, TNR here indicates true negative rate. However, when a dataset is class imbalanced, BA will have an implicit bias against the minority class. Therefore, we introduce Difference of Balanced Accuracy (DBA) as a way to measure the difference in a model’s performance across groups defined by a sensitive label while accounting for class imbalance, formally: $\mathrm{BA} = \frac{1}{2}|(\mathrm{TPR}_{s=1} + \mathrm{TNR}_{s=1}) - (\mathrm{TPR}_{s=0} + \mathrm{TNR}_{s=0})|$. 

\begin{figure*}[t]
	\centering 
	\includegraphics[ width=0.95\linewidth]{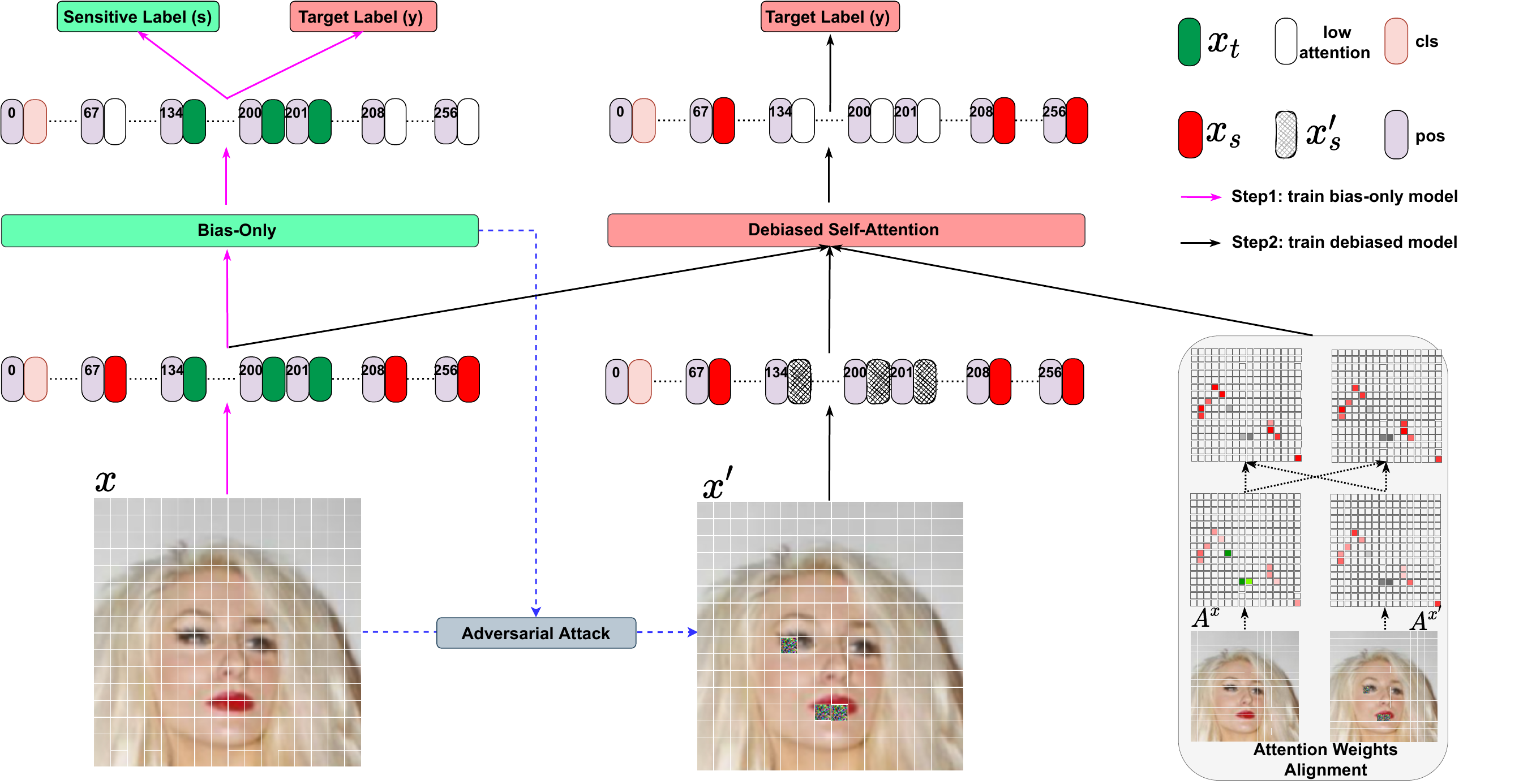}
	\caption{The DSA framework. The target label is \emph{Hair Color} and the sensitive label is \emph{Gender}. The bias-only model is first trained to learn the {\it spurious features} (the green patches) for predicting sensitive label $s$ but not to learn the real features (the red patches) with an adversarial objective. The adversarial attack is then applied against the bias-only model to generate the adversarial examples $x^\prime$, by perturbing the spurious features (the grid shadow patches) of the original inputs $x$ (see Section \ref{Sec: AT}). Finally, both $x$ and $x^\prime$ are used to train a fairness-aware ViT with an attention weights alignment objective (see Eq. \eqref{eq:celoss}) and learn the {\it real features} (the red patches) (see Sections \ref{Sec: AWA} and \ref{Sec: Oveall_Loss}).}
	\label{fig:architeture}
\end{figure*}

\noindent
\textbf{Problem Formulation:}
We consider a supervised classification task with training samples $\{x, y, s\} \sim p_{data}$, where $x \in \mathcal{X}$ is the input image, $y \in \mathcal{Y}$ is the target label and $s \in \mathcal{S}$ is the sensitive label. Some examples of $\mathcal{S}$ include gender, race, age, or other attributes that can identify a certain protected group. We assume that the sensitive label $s$ can only be used during the training phase, and are not accessible during the inference phase. Moreover, we assume that each input feature $\mathbf{x}$ can be split into two components, one with \textbf{spurious features} $\mathbf{x}_s$ that are highly correlated with the sensitive label $s$, and the rest $\mathbf{x}_t$ that are \textbf{real features} correlated with the target label $y$, i.e., $\mathbf{x} = (\mathbf{x}_s, \mathbf{x}_t)$. 

Deep learning models, including ViT, are trained using a large amount of data and learn patterns from the data to make predictions. However, if the training data is imbalanced or biased, the model can learn spurious patterns that reflect the biases in the data rather than the real patterns. This is a particular problem when there are spurious features $\mathbf{x}_s$ in the data that are highly correlated with the target label $y$. For example, if a dataset includes information about the race or gender of individuals, and these sensitive labels are highly correlated with certain outcomes, a model trained on that data may learn to use those spurious features to make predictions, a.k.a shortcut learning \cite{geirhos2020shortcut}. 

This motivates the proposal of our two-step hierarchical DSA framework for bias mitigation. In the first step, DSA localizes and masks the spurious features $\mathbf{x}_s$ from the input $\mathbf{x}$ to disentangle $\mathbf{x}_s$ from $\mathbf{x}_t$. This is accomplished by transforming the model prediction from $p(x) = p(y|\mathbf{x}_s, \mathbf{x}_t)$ to $p(x) \propto p(x^\prime) = p(y|\mathbf{x}^\prime_t)$, where $x^\prime$ is the sample constructed after masking the spurious features $\mathbf{x}_s$ from $\mathbf{x}$ via adversarial attacks. In the second step, DSA utilizes the original $x$ and the augmented data $x^\prime$ to train a ViT model, while at the same time satisfying certain fairness requirements (i.e., DP, EO, and DBA) concerning the sensitive label $s$. 

\section{Debiased Self-Attention (DSA) Framework}

\noindent
Our major motivation is to achieve fairness of ViT by mitigating the influence of spurious features (e.g., `red lip' and `eye shadow') on the prediction task ({e.g., `hair color'}). While CNNs also struggle with identifying spurious features, the ViT framework presents its own distinct challenge in locating these features directly from the input patches. To address this challenge, we propose a hierarchical framework, named DSA, in a two-step procedure as shown in Figure \ref{fig:architeture}:\\ \textbf{Step 1}: a {\em bias-only} model is first trained deliberately to {\it maximize} the usage of spurious features while {\it minimize} the usage of the real features for sensitive label prediction. Then, the adversarial examples, in which the spurious features are perturbed, are constructed via adversarial attacks against the {\em bias-only} model.\\ \textbf{Step 2}: a {\it debiased} model is trained with augmented adversarial examples and attention weights alignment aiming to mitigate the influence of spurious features on the prediction task to preserve the accuracy. 

\subsection{Training the Bias-only Model}

\label{Sec: BO_Model}
\noindent
Our initial goal for training the bias-only ViT is to build a model that exclusively learns spurious features while disregarding the real features. Recall that the input features consist of two components $\mathbf{x}=(\mathbf{x}_s, \mathbf{x}_t)$, where $\mathbf{x}_s$ and $\mathbf{x}_t$ denote the spurious and real features, respectively. Thus, our goal is to build the bias-only model, i.e., $f_B$, which only learns $\mathbf{x}_s$ but neglects $\mathbf{x}_t$ from the input features. The input image $x$ is first fed into the feature extractor of ViT $h$: $\mathcal{X} \rightarrow \mathbb{R}^k$, where $k$ is the feature dimension. Subsequently, the extracted features $h(\mathbf{x})$ are fed forward through both the sensitive label prediction head $f_s$: $\mathbb{R}^k \rightarrow \mathcal{S}$ and the target label prediction head $f_t$: $\mathbb{R}^k \rightarrow \mathcal{Y}$. $f_s$ here is trained to \textbf{minimize} the following Cross-Entropy (CE) loss: $\mathcal{L}_S(x,s) = \mathcal{L}_{\mathrm{CE}}(f_s(h(\mathbf{x}),s))$.
While the objective of training $f_t$ is to \textbf{maximize}: $\mathcal{L}_T(x,y) = \mathcal{L}_{\mathrm{CE}}(f_t(h(\mathbf{x}),y))$,
The two prediction heads $f_s$ and $f_t$ of the bias-only model $f_B$ are jointly trained in an adversarial training strategy as: $\mathcal{L}_B(x,s,y) = \mathcal{L}_S(x,s) - \mathcal{L}_T(x,y)$.
With this objective, $f_s$ is trained to minimize the CE loss to correctly predict $s$. While $f_t$ is trained to maximize the CE loss to refrain $f_t$ from predicting $y$.

In practice, $h$, $f_t$, and $f_s$ in the bias-only model $f_B$ are trained jointly with both adversarial strategy \cite{zhang2018mitigating} and gradient reversal technique \cite{ganin2015unsupervised}. Early in the learning, $f_s \odot h$ is rapidly trained to predict $s$ using spurious features. Then $f_t$ learns to refrain from predicting $y$, and $h$ learns to extract spurious features that are independent of $y$. At the end of the training, $f_B$ performs poorly in predicting the target label $y$ yet performs well in predicting the sensitive label $s$. It is not due to the divergence but due to the feature extractor $h$ {\bf unlearns} the real features. As such, $h(\mathbf{x})$ extracts largely spurious features instead of real features for the subsequent debiasing of ViT in Step 2.

We illustrate this idea using the example in the left panel of Figure \ref{fig:architeture}. We consider the \emph{Hair Color} prediction task with \emph{Gender} bias. The bias-only model $f_B$ mainly relies on the spurious features, like `eye shadow' and/or `red lips', to predict the sensitive label $s$ (e.g., \emph{Gender}), while at the same time paying nearly no attention to the real features, i.e., `hair', to predict the target label $y$ (e.g., \emph{Hair Color}).

\subsection{Adversarial Attack Against the Bias-only Model}

\label{Sec: AT}
\noindent
After building the bias-only model $f_B$, DSA uses adversarial attacks to construct augmented adversarial examples in which the spurious features are localized and perturbed as shown in Figure \ref{fig:architeture}. Since ViTs and CNNs have different architectures and input formats, traditional adversarial attacks, which are originally designed for CNNs, are not effective against ViTs \cite{fu2022patch}. Inspired by the idea of Patch-Fool \cite{fu2022patch}, we design a novel adversarial attack method aiming to localize and perturb the spurious features in the inputs and retain the real features at the same time. Specifically, our attack method is designed to compromise the self-attention mechanism in the pre-trained bias-only model by attacking its basic component (i.e., a single patch) with a series of attention-aware optimization techniques. In more details, given $\mathcal{L}_S(x,s)$, $\mathcal{L}_T(x,y)$, and a sequence of input image patches $\mathbf{X} = [\mathbf{x_1}, \cdots, \mathbf{x_p}, \cdots, \mathbf{x_n}]^\mathrm{T} \in  \mathbb{R}^{n \times d}$ with its associated sensitive label $s$, and target label $y$, the objective of the adversarial attack algorithm is
\begin{align}
    \underset{1 \leq p \leq n, \mathbf{E} \in \mathbb{R}^{n \times d}}{\arg\max} \mathcal{L}_S(\mathbf{X} + \mathds{1} \odot \mathbf{E}, s)\  + \underset{1 \leq p \leq n, \mathbf{E} \in \mathbb{R}^{n \times d}}{\arg\min} \mathcal{L}_T(\mathbf{X} + \mathds{1} \odot \mathbf{E}, y),
\end{align}
where $\mathbf{E}$ denotes the adversarial perturbation; $\mathds{1} \in \mathbb{R}^n$ is the identifying one-hot vector demonstrating whether current $p$-th patch is selected or not. $\odot$ here represents the element-wise multiplication of two matrices. 

This adversarial attack algorithm proceeds by (1) selecting the adversarial patch $p$ and (2) optimizing the corresponding adversarial attack $\mathbf{E}$.

\noindent
\textbf{Selection of $p$}: For encoder blocks in the ViT, we define: $t_j^{(l)} = \sum_{h,i} a_j^{(l,h,i)}$ to measure the importance of the $j$-th patch in the $l$-th block based on its contributions to other patches in the self-attention calculation, where $\mathbf{a}^{(l,h,i)} = [a_1^{(l,h,i)}, \cdots, a_n^{(l,h,i)}]$ denotes the attention distribution for the $i^{\text{th}}$ patch of the $h^{\text{th}}$ head in the $l^{\text{th}}$ block. Our motivation is to localize the most influential patch $p$ according to the predicted sensitive label $s$ but with the least impact on predicting target label $y$. Here, we derive the top $k$, which is a tunable hyper-parameter, important patches from ${\arg\max} ~ t_j^{(l)}$.

\noindent
\textbf{Optimization of $\mathbf{E}$}: Given the selected adversarial patch index $p$ from the previous step, an attention-aware loss is applied for the $l^{\text{th}}$ block as $\mathcal{L}_\mathrm{Attn} = \sum_{h,i} a_p^{(l,h,i)}$.  This loss is expected to be maximized so that the adversarial patch $p$, serving as the target adversarial patch, can attract more attention from other patches for effectively fooling ViTs. The perturbation $\mathbf{E}$ is then updated based on both the final sensitive label classification loss and a layer-wise attention-aware loss:
\begin{align}
  \mathcal{L}(\mathbf{X}^\prime,s,p) = \mathcal{L}_S(\mathbf{X}^\prime,s) -  \mathcal{L}_T(\mathbf{X}^\prime,y) + \alpha \underset{l}{\sum}\mathcal{L}_\mathrm{Attn}(\mathbf{X}^\prime,p),
\end{align}
where $\mathbf{X}^\prime \triangleq \mathbf{X} +\mathds{1} \odot \mathbf{E}$ and $\alpha$ is a weight hyper-parameter set to $0.5$ in the experiments. Moreover, PCGrad \cite{yu2020gradient} is adopted to avoid the gradient conflict of the two losses and $\mathbf{E}$ is updated using:
\begin{align}
  \delta_\mathbf{E} = \nabla_\mathbf{E}\mathcal{L}(\mathbf{X}^\prime,s,p) - \alpha\underset{l} {\sum}\beta_l\nabla_\mathbf{E}\mathcal{L}_S(\mathbf{X}^\prime,s),
\end{align}
\begin{align}
    \beta_l = \left\{
    \begin{aligned}
    &0,  \ \ \ \ \ \ \ \ \langle \nabla_\mathbf{E}\mathcal{L}_S(\mathbf{X}^\prime,s), \nabla_\mathbf{E}\mathcal{L}_\mathrm{Attn}(\mathbf{X}^\prime,p) \rangle > 0 \\
    &\frac{\langle \nabla_\mathbf{E}\mathcal{L}_S(\mathbf{X}^\prime,s), \nabla_\mathbf{E}\mathcal{L}_\mathrm{Attn}(\mathbf{X}^\prime,p) \rangle}{\|\nabla_\mathbf{E}\mathcal{L}_S(\mathbf{X}^\prime,s)\|^2},~~~ \mathrm{otherwise}.
    \end{aligned}
\right.
\end{align}
Following PGD \cite{madry2017towards}, we iteratively update $\mathbf{E}$ using an Adam optimizer: $\mathbf{E}^{t+1} = \mathbf{E}^t + \eta \cdot \mathrm{Adam}(\delta_{\mathbf{E}^t})$, where $\eta$ is the step-size for each update.

\subsection{Attention Weights Alignment}

\label{Sec: AWA}
After Step 1, the DSA framework generates the adversarial example $x^\prime$, whose top $k$ patches containing spurious features are perturbed through the adversarial attack. Here, besides using these adversarial examples as augmentation during training of the debiased ViT model, we also leverage them via attention weights alignment to further guide the model to pay more attention to the real features. This further allows more spurious features to be discovered and ignored by the self-attention mechanism in the ViT model as shown in the right panel of Figure \ref{fig:architeture}. 

In particular, we apply three different feature discrepancy metrics $D(\cdot,\cdot)$, i.e., Mean Squared Error (MSE), Kullback-Leibler Divergence (KL-Div), and Attention Transfer (AT), to evaluate the discrepancy between the attention weights $\mathbf{A}^{x}$ and $\mathbf{A}^{x^\prime}$ from the original sample $x$ and the adversarial example $x^\prime$, respectively. Formally:
\begin{align}
    &D_{\mathrm{MSE}}(\mathbf{A}^x,\mathbf{A}^{x^\prime}) = \frac{1}{2} \underset{j \in \mathcal{I}}{\sum} \|\mathbf{A}^x_j - \mathbf{A}^{x^\prime}_j\|_2, \\
    &D_{\mathrm{KL-Div}}(\mathbf{A}^x \| \mathbf{A}^{x^\prime}) = \underset{j \in \mathcal{I}}{\sum} \mathbf{A}^x_j \log \frac{\mathbf{A}^x_j}{\mathbf{A}^{x^\prime}_j}, \\
    \label{eq:D}
    &D_{\mathrm{AT}}(\mathbf{A}^x,\mathbf{A}^{x^\prime}) = \frac{1}{2} \underset{j \in \mathcal{I}}{\sum} \bigg\|\frac{\mathbf{A}^x_j}{\|\mathbf{A}^x_j\|_2} - \frac{\mathbf{A}^{x^\prime}_j}{\|\mathbf{A}_j^{x^\prime}\|_2}\bigg\|_2,
\end{align}
where $\mathcal{I}$ denotes the indices of all the adversarial examples and the original example attention weights pairs for which we perform the alignment. Finally, to incorporate the attention distributions of $\mathbf{A}^{x}$ and $\mathbf{A}^{x^\prime}$ in the objective, we add $\mathcal{L}_A = D(\mathbf{A}^{x}, \mathbf{A}^{x^\prime})$ as a regularizer in the overall training objective.

\subsection{Overall Training Objective}

\label{Sec: Oveall_Loss}
\noindent
Putting the above Steps 1 and 2 together, the overall objective for training the proposed debiased model is: 
\begin{align} \label{eq:celoss}
  \mathcal{L} = \lambda_1 \mathcal{L}_{CE}(x, y) + \lambda_2 \mathcal{L}_{CE}(x^\prime, y) + \lambda_3 \mathcal{L}_A,
\end{align} 
where $\mathcal{L}_{CE}$ denotes the standard Cross-Entropy loss. $\lambda_1, \lambda_2$, and $\lambda_3$ are three tunable weights for controlling the fairness-utility trade-off. We have
conducted a comprehensive examination of how varying these weights would impact model performance. We fine-tuned these weights across various values with the results shown in Table \ref{tab:tuning}. 

\section{Experimental Settings}

\noindent
\textbf{Datasets:}
We evaluate the DSA framework on three CV datasets widely used in the fairness research community, namely, Waterbirds \cite{sagawa2019distributionally}, CelebA \cite{liu2015deep}, and bFFHQ \cite{kim2021biaswap}. It is important to note that all the datasets are derived from real-world data and offer a significant variety of challenges to real-world applications, such as facial recognition.
% The promising performance, as shown in Table \ref{tab:mainresults}, on these benchmarks, demonstrates the solid foundation, facilitating the extension and adaptation of DSA to real-world applications.}
Waterbirds dataset contains spurious correlation between the background features $\mathcal{S}$ = \{Water, Land\} and target label $\mathcal{Y}$ = \{Waterbird, Landbird\}. The spurious correlation is injected by pairing waterbirds with the water background and land birds with the land background more frequently, as compared to other combinations. CelebA dataset contains 200k celebrity face images with annotations for 40 binary attributes. We present the results on the settings following \cite{sudhakar2023mitigating}, where Hair Color (gray or not gray) is the target label the model is trained to predict and Gender is the sensitive label over which we wish the model to be unbiased. We select Hair Color for two reasons: it shows a significant performance difference between male and female groups (as shown in Table \ref{tab:mainresults}), indicating bias, and hair has less gender overlap in facial images (as shown in Figure \ref{fig:grayhair_example}). Thus, the adversarial attack targets gender-related features without obscuring real features like hair. bFFHQ dataset has Age as a target label and Gender as a correlated bias. The images in the bFFHQ dataset include the dominant number of young women (i.e., aged 10-29) and old men (i.e., aged 40-59) in the training data. 
% We provide more details of these datasets in the Appendix. 

\noindent
\textbf{Implementation Details:}
We train the ViT-S/16 models from scratch for each prediction task. The ViT-S/16 model consists of 196 patches (each representing a 16x16 sub-image), 1 class token patch, 12 transformer encoder layers, and 8 attention heads. We flatten and project each patch into a 64-dimensional vector and add positional embeddings. The embedded patches are fed into the ViT encoder. After the ViT encoder processes the patch embeddings, the class token patch is fed into 2 fully-connected layers (with a hidden state size of 256) and a sigmoid layer to produce a single normalized output score (since we deal with binary classification). We train the ViT models using momentum Stochastic Gradient Descent (SGD) with a momentum parameter of 0.9 and an initial learning rate of 3e-2 for 20 epochs. We use a batch size of 32, gradient clipping at global norm $1$, and a cosine decay learning rate schedule with a linear warmup following \cite{ghosal2022vision}. We select the model with the best accuracies on the validation sets.

\noindent
\textbf{Baselines:}
Since our DSA is an in-processing debiasing method, we have chosen the following debiasing algorithms from the in-processing category as baselines for a fair performance evaluation. To our knowledge, besides the proposed DSA and the in-house baseline AM methods, TADeT is the only third-party fairness-aware algorithm tailor-made for ViT, while all the others are originally designed for CNNs. We consider the following baselines:
\noindent
\textbf{Vanilla \cite{dosovitskiy2020image}:} The ViT models are only trained with CE loss for target prediction. 
\noindent
Mitigating Bias in ViT via Target Alignment \textbf{(TADeT)} \cite{sudhakar2023mitigating} uses a targeted alignment strategy for debiasing ViT that aims to identify and remove bias primarily from query matrix features.
\noindent
Maximum Mean Discrepancy \textbf{(MMD)} \cite{long2015learning} calculates the mean of penultimate layer feature activation values for each sensitive label setting and then minimizes their $L_2$ distance.
\noindent
MMD-based Fair Distillation \textbf{(MFD)} \cite{jung2021fair} adds an MMD-based regularizer that utilizes the group-indistinguishable predictive features from the teacher model while discouraging the student model from discriminating against any protected group.
\noindent
Domain Adversarial Neural Network \textbf{(DANN)} \cite{ganin2015unsupervised} employs a sensitive label adversary learned on top of the penultimate layer activation. The adversarial head consists of two linear layers in the same dimension as the class token, followed by a sigmoid function. 
\noindent
Learning Adversarially Fair and Transferable Representation \textbf{(LAFTR)} \cite{madras2018learning} trains a model with a modified adversarial objective that attempts to meet the fairness criterion, e.g., EO. This objective is implemented by minimizing the average absolute difference on each task. 
\noindent
Attention Masking \textbf{(AM)}: The self-attention mechanism is critical in ViT as it provides important weights for extracting visual features. We propose the AM method as a {\it home run} that directly masks the top-$k$ patches with the highest attention scores for the bias-only model.

\section{Results and Discussion}

% In this Section, we report the results of fairness and accuracy evaluations and the ablation study. In Appendix, more experimental results are reported, including the effects of model size and patch size, the impact of several tunable hyper-parameters, results with different $D$ in the regularizer $\mathcal{L}_A$, and some other qualitative evaluations. 

\begin{table*}[t]
\footnotesize
\centering
\caption{Fairness and accuracy evaluation for our methods and other baseline methods over different combinations of the target label ($y$) and the sensitive label ($s$) on the three datasets. For DSA, we use $\mathcal{L}_A = D_{AT}$ as the attention weight alignment regularizer. The $k$ values of AM and DSA are both set as 3. The best results are bold-faced. \label{tab:mainresults}}
\resizebox{\textwidth}{!} {
\begin{tabular}{l|ccccc|ccccc|ccccc}
\hline
\multirow{2}{*}{Methods}      & \multicolumn{5}{c|}{Waterbirds\ \ $Y$: Bird Type \ \ \ $S$: Background}                                & \multicolumn{5}{c|}{bFFHQ\ \ $Y$: Age \ \ \ $S$: Gender}                                         & \multicolumn{5}{c}{CelebA\ \ $Y$: Hair Color \ \ \  $S$: Gender}                             \\ \cline{2-16} 
& EO$\downarrow$ & DP$\downarrow$ & DBA$\downarrow$ & BA(\%)$\uparrow$ & Acc(\%)$\uparrow$
& EO$\downarrow$ & DP$\downarrow$ & DBA$\downarrow$ & BA(\%)$\uparrow$ & Acc(\%)$\uparrow$
& EO$\downarrow$ & DP$\downarrow$ & DBA$\downarrow$ & BA(\%)$\uparrow$ & Acc(\%)$\uparrow$ \\
\hline
Vanilla    & 0.0209 & 0.0211 & 0.0841 & 60.24 & 62.36 & 0.3214 & 0.3410 & 0.1021 & 68.64 & 75.93 & 0.2763 & 0.3185 & 0.0422 & 81.84 & 90.25    \\
TADeT    & 0.0424 & 0.0266 & 0.0747 & \textbf{64.14} & 69.05 & 0.3189 & 0.3318 & 0.0944 & 69.06 & 77.05 & 0.2850 & 0.2422 & 0.0427 & 81.27 & 90.23    \\
MMD    & 0.0617 & 0.0380 & 0.0767 & 63.52 & 67.81 & 0.3041 & \textbf{0.2774} & \textbf{0.0847} & 70.59 & 76.59 & 0.3135 & 0.3023 & 0.0112 & 80.77 & 90.02    \\
MFD    & 0.0386 & 0.0297 & 0.0736 & 63.01 & 67.36 & 0.2922 & 0.3135 & 0.0912 & 68.97 & 77.63 & 0.2812 & 0.3049 & 0.0237 & 81.74 & 90.41    \\
DANN    & 0.0337 & 0.0238 & 0.0951 & 58.64 & 60.04 & 0.3067 & 0.3274 & 0.1141 & 69.87 & 76.75 & 0.2720 & 0.2586 & 0.0134 & 82.15 & 90.69    \\
LAFTRE    & 0.0822 & 0.0415 & 0.0814 & 61.36 & 64.80 & 0.2936 & 0.3075 & 0.0961 & 70.05 & 76.67 & 0.3094 & 0.2682 & 0.0411 & 79.87 & 89.32   \\ \hline
AM    & 0.0447 & 0.0332 & 0.0872 & 59.98 & 61.70 & 0.2874 & 0.2978 & 0.1021 & 70.91 & 78.76 & 0.2877 & 0.2621 & 0.0256 & 81.51 & 90.35   \\
{\bf DSA}  & \textbf{0.0185} & \textbf{0.0113} & \textbf{0.0709} & 63.87 & \textbf{69.58} & 0.\textbf{2651} & 0.2856 & 0.0879 & \textbf{71.37} & \textbf{78.82} & \textbf{0.2558} & \textbf{0.2337} & \textbf{0.0031} & \textbf{82.92} & \textbf{90.95}   \\
\hline
\end{tabular}
}
\end{table*}

\subsection{Fairness and Accuracy Evaluations}

\noindent
We report the fairness and accuracy performance of the three datasets in Table \ref{tab:mainresults} with the following observations. First, DSA outperforms the baselines on most evaluation metrics improving the ViT fairness with lower EO, DP, and DBA while maintaining higher accuracy in terms of BA and ACC. 

Second, several baseline methods (e.g., MMD, MFD, and DANN) that have shown strong performance with CNN models, do not even outperform the vanilla model on some fairness metrics (e.g., EO), particularly on the waterbird dataset. 
This may be attributed to the intrinsic differences between their architectural designs and feature learning mechanisms. ViTs are designed with transformers to capture the global context and long-range dependencies, while CNNs excel at capturing local patterns and spatial hierarchies. Additionally, ViTs employ self-attention mechanisms to capture relationships between input image patches, whereas CNNs rely on local receptive fields and shared weights for feature extraction. As such, these baseline methods (designed for the CNNs) are not transferable for bias mitigation with the ViT models. 

Third, we note the in-house baseline method AM is also designed by blinding the spurious features in the input based on only the attention weights of the bias-only model. However, several works \cite{serrano2019attention,jain2019attention} have questioned whether highly attentive inputs would significantly impact the model outputs. Since the self-attention mechanism involves the computation of queries, keys, and values, reducing it only to the derived attention weights (inner products of queries and keys) can be insufficient to capture the importance of the features. Hence, the {\it home run} AM method fails to achieve comparable performance with the proposed DSA method.

\begin{table}[t]
\footnotesize
\centering
\caption{Ablation study of the three training objectives on the CelebA and Waterbirds datasets. The best results are bold-faced. `w/o' represents without. \label{tab:ablation}}
\begin{tabular}{l|ccccc|ccccc}
\hline
\multirow{2}{*}{Models}      & \multicolumn{5}{c}{CelebA $Y$: Hair Color\ \ $S$: Gender}  & \multicolumn{5}{|c}{Waterbirds $Y$: Bird Type\ \ $S$: Background} \\ \cline{2-11} 
& EO$\downarrow$ & DP$\downarrow$ & DBA$\downarrow$ & BA$\uparrow$ & Acc$\uparrow$ 
& EO$\downarrow$ & DP$\downarrow$ & DBA$\downarrow$ & BA$\uparrow$ & Acc$\uparrow$\\
\hline
$\mathcal{L}$(all)     & \textbf{0.2558} & \textbf{0.2337} & \textbf{0.0031} & \textbf{82.92} & \textbf{90.95} & \textbf{0.0185} & \textbf{0.0113} & \textbf{0.0709} & 63.87 & \textbf{69.58} \\
w/o $\mathcal{L}_{CE}(x,y)$    & 0.2754 & 0.2541 & 0.0175 & 81.21 &  88.32  & 0.0249 & 0.0275 & 0.0843& 61.12 & 66.43\\
w/o $\mathcal{L}_{CE}(x^\prime,y)$  & 0.2641  & 0.2503 & 0.0129 & 80.65 & 88.54   & 0.0265 & 0.0234 & 0.0914 & 62.58 & 67.21\\
w/o $\mathcal{L}_{A}$    & 0.2934 & 0.2865 & 0.0206 & 81.54 & 89.91   & 0.0285 & 0.0309 & 0.0806 & \textbf{64.03} & 67.85\\
\hline 
\end{tabular}
\end{table}

\subsection{Ablating DSA}
\label{Sec: Abalation}

\noindent
The objective of DSA contains three components for bias mitigation as shown in Eq.\eqref{eq:celoss}. We conduct an ablation study on the CelebA and Waterbird datasets to analyze their individual contributions and report the results in Table \ref{tab:ablation}. We summarize our major findings. First, all of the components contribute towards improved prediction and fairness performance across all metrics as shown in the row named $\mathcal{L}$(all). Second, both the CE losses, i.e., $\mathcal{L}_{CE}(x,y)$ and $\mathcal{L}_{CE}(x^\prime,y)$, in Eq.\eqref{eq:celoss} are critical in preventing DSA from compromising the prediction performance; otherwise, the accuracies drop from 90.95 to 88.32/88.54 on the CelebA dataset and 69.58 to 66.43/67.21 on the Waterbirds dataset, respectively. Third, the regularizer $\mathcal{L}_{A}$ contributes the most to effective debiasing in ViTs, as evidenced by the higher fairness measures in Table \ref{tab:ablation}.

\begin{table}[t]
\footnotesize
\centering
\caption{Performance evaluation with different ViT models (i.e., ViT-B (B), ViT-S (S), and DeiT (D)) and patch sizes (i.e., 16 and 32). VA denotes the vanilla ViT. \label{tab:model}}
\begin{tabular}{cc|ccccc}
\hline
\multicolumn{2}{c}{\multirow{2}{*}{Model}} & \multicolumn{5}{|c}{$Y$: Y:\ \ \ $S$: Gender} \\
\cline{3-7} 
& & EO$\downarrow$ & DP$\downarrow$ & DBA$\downarrow$ & BA$\uparrow$ & ACC$\uparrow$ \\
\hline
\multirow{2}{*}{B/16}  & VA & 0.2984 & 0.2841 & 0.0142 & 81.95 & 91.05  \\
                       & DSA & \textbf{0.2424} & \textbf{0.2205} & 0.0081 & \textbf{83.42} & \textbf{91.24}  \\
\multirow{2}{*}{S/16}  & VA & 0.2763 & 0.3185 & 0.0422 & 81.84 & 90.25  \\
                       & DSA & 0.2558 & 0.2337 & \textbf{0.0031} & 82.92 & 90.95  \\
\hline
\multirow{2}{*}{B/32}  & VA  & 0.2982 & 0.2976 & 0.0205 & 81.11 & 90.16  \\
                       & DSA  & \textbf{0.2629} & \textbf{0.2520} & 0.0109 & \textbf{82.73} & \textbf{91.03}  \\
\multirow{2}{*}{S/32}  & VA  & 0.3014 & 0.3213 & 0.0198 & 80.64 & 89.18  \\
                       & DSA  & 0.2935 & 0.3165 & \textbf{0.0086} & 80.86 & 89.45  \\ \hline 
\multirow{2}{*}{D/16}     & VA & 0.0743 & 0.0942 & 0.0268 & 94.04 & 96.57  \\
                       & DSA & \textbf{0.0674} & \textbf{0.0654} & \textbf{0.0088} & \textbf{94.33} & \textbf{96.61}  \\
\hline 
\end{tabular}
\end{table}

\subsection{Effects on DSA Performance}

\noindent
\textbf{Effect of ViT Model Size and Patch Size:}
We further examine the effect of ViT architecture, model size, and patch size on DSA. The ViT-B model is larger than the ViT-S model, which has 12 self-attention heads in each block and 256 hidden state sizes in the two fully-connected layers. Each patch is flattened and projected into a vector of 768 dimensions. DeiT \cite{touvron2021training} shares a similar architecture with ViT but introduces a distillation token that interacts with class and patch tokens. We draw several conclusions from Table \ref{tab:model}. First, the larger ViT-B models outperform the smaller ViT-S on most of the fairness and accuracy metrics, demonstrating better feature learning capabilities with higher feature dimensions and more self-attention heads. Second, a smaller patch size performs better on both fairness and accuracy measurements because small patches enable extracting more fine-grained features. Lastly, our DSA performs effectively on both ViT architectures, demonstrating its strong debiasing capabilities and generalizability.

\noindent
\textbf{Effect of Discrepancy Metrics:} 
Since three different metrics $D$, i.e., MSE, KL-Div, and AT, are applied to evaluate the discrepancy between the attention weights $\mathbf{A}^{x}$ and $\mathbf{A}^{x^\prime}$, we report the effect of these discrepancy metrics in Table \ref{tab:tuning}. Although the differences between these discrepancy metrics are relatively small, AT clearly achieves the best performance, especially on the fairness metrics, i,e, EO, DP, and DBA. Since AT can capture the most significant differences between $\mathbf{A}^{x}$ and $\mathbf{A}^{x^\prime}$ as shown in Eq. \eqref{eq:D}, the regularizer $\mathcal{L}_A$ is more efficient to minimize their differences. In the following results, we use $\mathcal{L}_A = D_{AT}$ in DSA as the attention weight alignment regularize.

\begin{table}[t]
\footnotesize
\centering
\caption{Evaluations with different tunable hyper-parameters and discrepancy metrics on the CelebA dataset. $\lambda_1$, $\lambda_2$, $\lambda_3$ are the coefficient weights in the objective function Eq.\eqref{eq:celoss}. $k$ represents the different number of perturbed patches during the adversarial attack. \label{tab:tuning}}
\begin{tabular}{cc|ccccc}
\hline
\multirow{2}{*}{Hyper-parameters} & \multirow{2}{*}{Values}  & \multicolumn{5}{c}{$Y$: Hair Color\ \ $S$: Gender}   \\ 

& & EO$\downarrow$ & DP$\downarrow$ & DBA$\downarrow$ & BA$\uparrow$ & Acc\%$\uparrow$  \\
\hline
\multirow{3}{*}{$\lambda_1$, $\lambda_2$, $\lambda_3$} & 1.0, 0.5, 0.5 & 0.2843 & 0.2675 & 0.0125 & 81.45 & \textbf{91.12}   \\
& 0.5, 1.0, 0.5 & 0.2633 & 0.2578 & 0.0106 & 81.32 & 89.26  \\
& 1.0, 1.0, 0.5 & \textbf{0.2558} & \textbf{0.2337} & \textbf{0.0031} & \textbf{82.92} & 90.95  \\
\hline
\multirow{3}{*}{k} & 1   & 0.2946 & 0.3075 & 0.0110 & 81.77 & 90.61    \\
& 3   & \textbf{0.2558} & \textbf{0.2337} & \textbf{0.0031} & \textbf{82.92} & \textbf{90.95}    \\
& 5   & 0.2776 & 0.2560 & 0.0216 & 81.91 & 88.55    \\
\hline 
\multirow{3}{*}{Discrepancy  Metrics} & MSE & 0.2706 & 0.2488 & 0.0136 & 82.07 & 90.13   \\
& KL-Div & 0.2608 & 0.2467 & 0.0106 & \textbf{83.26} & 89.48  \\
& AT & \textbf{0.2558} & \textbf{0.2337} & \textbf{0.0031} & 82.92 & \textbf{90.95}  \\
\hline 
\end{tabular}
\end{table}

\noindent
\textbf{Effect of Tunable Hyper-parameters:} 
There are several tunable hyper-parameters in the proposed DSA framework, including the various coefficient weights in the objective function and the number of masked patches learned during the adversarial attack. We tune the three coefficient weights in the objective function Eq. \eqref{eq:celoss} to identify the best-performing model as shown in Table \ref{tab:tuning}. To improve model performance, we believe that these coefficient weights should be carefully tuned and selected under different settings and datasets. In our experiments, the ViT model with $k = 3$ patches achieves the best performance among all compared metrics in most settings. If we perturb only one patch out of all the input patches, some sensitive attributes may not be localized and masked. On the contrary, perturbing excessive patches (e.g., 5 patches) would increase the risk of masking the related attributes to the target task, resulting in worse prediction performance. For example, the Acc drops from 90.95 to 88.55 in the setting of (Hair Color, Gender) with 5 perturbed patches, as shown in Table \ref{tab:tuning}. 

\subsection{Performance with Post-processing Method} 

We further implement RTO \cite{alabdulmohsin2021near}, which is a scalable post-processing algorithm for debiasing. Clearly, implementing RTO results in marginally reduced statistical parity (SP), as shown in Table \ref{tab:rto}, showcasing its effectiveness in promoting fairer outcomes. In comparison to VA, DSA achieves higher Acc and lower SP both with and without RTO, demonstrating its effectiveness in debiasing ViT without compromising performance.

\begin{table}[t]
\footnotesize
\centering
\caption{Fairness and accuracy evaluation using DSA and vanilla ViT with and without RTO on CelebA. `w/' and `w/o’ represent with and without, respectively. \label{tab:rto}}
\begin{tabular}{cc|cccc}
\hline
Model                    & RTO & Acc (\%)$\uparrow$   & $\Delta$ Acc (\%) & SP$\downarrow$     & $\Delta$ SP (\%) \\ \hline
\multirow{2}{*}{VA} & w/  & 51.80 & \multirow{2}{*}{42.10}    & 0.4103 & \multirow{2}{*}{32.26}   \\ 
                         & w/o & 89.47 &                           & 0.6057 &                          \\ \hline
\multirow{2}{*}{DSA}     & w/  & 51.69 & \multirow{2}{*}{42.72}    & \textbf{0.4002} & \multirow{2}{*}{32.70}   \\ 
                         & w/o & \textbf{90.25} &                           & 0.5947 &                          \\ \hline
\end{tabular}
\end{table}

\subsection{Qualitative Evaluations} 

\begin{figure}[t]
    \centering
    \includegraphics[width = 0.45\linewidth]{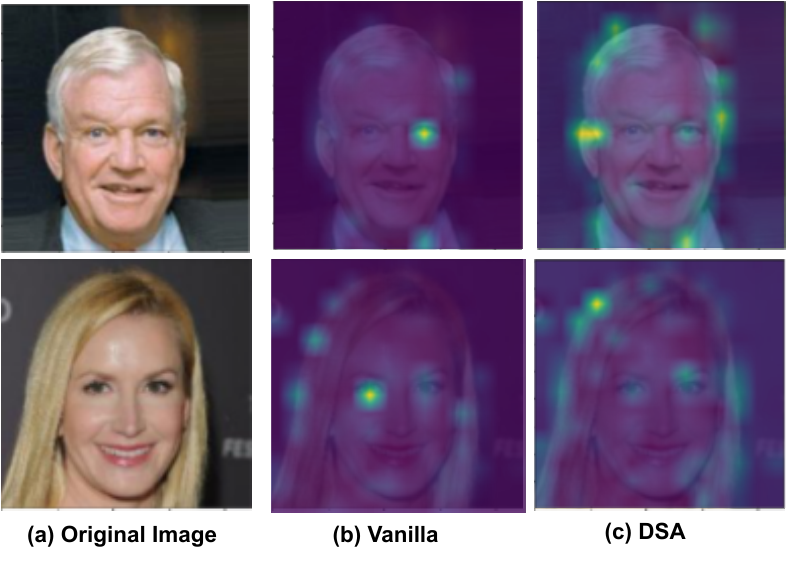}
    \caption{Qualitative evaluation. {\bf Y}: \emph{Hair Color}, {\bf S}: \emph{Gender}. \label{fig:grayhair_example}}
\end{figure}

\noindent
Figure \ref{fig:grayhair_example} further demonstrates the effectiveness of DSA. We note that the distribution of the attention weights for the vanilla ViT model largely focuses on the spurious feature, e.g., `eye shadow'. This demonstrates that the vanilla ViT model simply leverages the spurious features to predict the target label. On the contrary, DSA reduces the attention to these spurious features but pays more attention to the real features, e.g., `hair'. 

\section{Conclusion}

\noindent
In this work, we proposed a novel hierarchical fairness-aware ViT training framework named DSA for bias mitigation in both the training set and the learning algorithm while maintaining prediction performance. DSA is designed to eliminate spurious features in the data through adversarial attacks on the bias-only model, while also retaining the real features through an attention weights alignment regularizer. The quantitative and qualitative experimental results demonstrate the effectiveness of DSA for bias mitigation without compromising prediction performance.

\clearpage  % TODO REVIEW/FINAL: This \clearpage needs to be removed from both review and camera-ready versions.

%\section*{Acknowledgements}
%This work is supported by the National Science Foundation under grant ITE-2235225 and IIS-2211897. 


\bibliographystyle{splncs04}
\bibliography{main}
\end{document}